\documentclass[letterpaper]{article} 
\usepackage{aaai24}  
\usepackage{times}  
\usepackage{helvet}  
\usepackage{courier}  
\usepackage[hyphens]{url}  
\usepackage{graphicx} 
\usepackage{natbib}  
\usepackage{caption} 
\usepackage{algorithm}
\usepackage{algorithmic}
\usepackage{amsmath}

\usepackage{newfloat}
\usepackage{listings}
\usepackage{xcolor}
\usepackage{booktabs} 
\DeclareCaptionStyle{ruled}{labelfont=normalfont,labelsep=colon,strut=off} 
\floatstyle{ruled}
\newfloat{listing}{tb}{lst}{}
\floatname{listing}{Listing}
\usepackage{amssymb}
\usepackage{amsmath}
\usepackage{amsfonts}
\usepackage{booktabs}     
\usepackage{tabularx} 

\title{Offline Model-Based Optimization via Policy-Guided Gradient Search}
\author{Yassine Chemingui, 
        Aryan Deshwal,
        Trong Nghia Hoang,
        Janardhan Rao Doppa \\}

\affiliations{
School of EECS, Washington State University\\
{\{yassine.chemingui, aryan.deshwal, trongnghia.hoang,
jana.doppa\}@wsu.edu}
}

\begin{document}

\maketitle

\begin{abstract}
Offline optimization is an emerging problem in many experimental engineering domains including protein, drug or aircraft design, where online experimentation to collect evaluation data is too expensive or dangerous. To avoid that, one has to optimize an unknown function given only its offline evaluation at a fixed set of inputs. A naive solution to this problem is to learn a surrogate model of the unknown function and optimize this surrogate instead. However, such a naive optimizer is prone to erroneous overestimation of the surrogate (possibly due to over-fitting on a biased sample of function evaluation) on inputs outside the offline dataset. Prior approaches addressing this challenge have primarily focused on learning robust surrogate models. However, their search strategies are derived from the surrogate model rather than the actual offline data. To fill this important gap, we introduce a new {\em learning-to-search} perspective for offline optimization by reformulating it as an offline reinforcement learning problem. Our proposed policy-guided gradient search approach explicitly learns the best policy for a given surrogate model created from the offline data. Our empirical results on multiple benchmarks demonstrate that the learned optimization policy can be combined with existing offline surrogates to significantly improve the optimization performance.
\end{abstract}

\section*{Introduction}
\label{sec:intro}

Many science and engineering applications involve optimizing expensive-to-evaluate black-box functions over complex design spaces \cite{wang2023scientific}. Some prototypical examples include design optimization over input space of candidate proteins \cite{gao2020deep}, molecules \cite{deshwal2021bayesian}, drugs \cite{schneider2020rethinking}, hardware architectures \cite{huang2021machine,MOOS}, and superconducting materials \cite{ClaraNeurIPs20}. In such applications, evaluating a candidate input involves performing a lab experiment or expensive computational simulation. These problems are referred to as expensive black-box optimization (BBO). One standard framework to solve expensive BBO problems is Bayesian optimization where we iteratively query the black-box function's evaluation for inputs recommended by a surrogate model, whose accuracy is continuously improved via learning from such input-output pairs \cite{shahriari2015taking,eriksson2019scalable,LADDER}.

However, in some real-world scenarios where the overhead and cost of setting up experiments are prohibitively expensive (e.g., wet lab experiments that often require expensive materials and equipment), it becomes impractical to consider black-box optimization in the online setting. Instead, a more practical setting is to assume access to an existing database of previously collected input-output pairs, and consider solving this problem in an offline manner. This problem setting is termed \emph{offline model-based BBO} and was first introduced in the recent work of \cite{MINS,TrabuccoICML21}. The goal of optimization method is to leverage this offline training data to uncover optimal designs from the given input space. 
There are two main interrelated challenges for effectively solving the offline optimization problem. First, how to learn surrogate models from the offline dataset which are robust on inputs outside the offline dataset. Second, how to create effective search strategies beyond the local neighborhood of training data to find high-quality inputs. These challenges are often amplified due to the complexity of optimization problems including high-dimensional search spaces, highly sensitive objective functions where similar inputs might not induce similar outputs, and sparsity of the available data. 

Prior work on offline optimization (see Section~\ref{sec:related}) can be grouped into two broad categories based on the algorithmic design choices to address these two challenges. The first family
learns a generative model of the input distribution along side the surrogate model to characterize a trust region from which high-performing inputs can be sampled directly \cite{ClaraNeurIPs20, BrookeICML19}. However, generative models cannot be robustly learned on high-dimensional search spaces with sparse training data and domain knowledge required to create valid trust regions may not be available. The second family learns a surrogate model from the offline data which is then optimized directly using gradient updates  \cite{TrabuccoICML21, YuNeurIPs21,FuICLR21}. Conservative regularizers are typically designed to avoid overestimation for inputs which are far away from the offline training data. 
However, surrogate models, such as deep neural networks, can be non-smooth  which could result in highly sub-optimal solutions for a fixed gradient search strategy. This motivates the following question: {\em How can we learn policies to effectively guide the gradient-based search process in input regions outside the offline data?} 

This paper answers the above question by introducing a new learning-to-search perspective and a principled approach referred to as {\em Policy-guided Gradient Search (PGS)} for solving offline optimization problems. PGS is inspired by the prior successes of integrating learning into search procedures \cite{daume2009search, hc-search} and learned optimizers \cite{flennerhag2019meta}. The key idea behind PGS is to reformulate the step-size in the standard gradient update into a direction vector. Such direction vector can be viewed as an output $\alpha$ = $\pi(x_k)$ of a guiding policy $\pi$ to direct the search space exploration from $x_k$ in the direction of the high-performing input regions. PGS can be seen as a mechanism to correct the surrogate model’s gradients with respect to Oracle search. We formulate the policy learning task as an instance of offline reinforcement learning (RL) problem and provide an effective strategy for synthesizing trajectories from the offline dataset which is critical for creating high-performing policies. This reduction approach allows us to leverage the large body of existing work on offline RL to effectively solve offline optimization problems.  Hyper-parameter selection is a challenging task for offline optimization and none of the prior methods provide any procedure. To address this challenge, we propose an approach referred to as offline state estimation via latent embeddings using only the offline dataset. Empirical evaluation on diverse optimization tasks from the design-bench benchmark \cite{TrabuccoArXiv22} demonstrates that PGS performs better than prior methods on many tasks and ablation analysis shows the benefits of our algorithm design choices.

\vspace{1.0ex}

\noindent {\bf Contributions:} The key contribution of this paper is the development and evaluation of the PGS approach for solving offline optimization problems. Specific contributions are:

\begin{itemize}

\item Learning-to-search formulation to guide gradient search and reducing policy learning to offline RL.

\item Trajectory synthesis for offline RL based policy learning and hyperparameter tuning using trajectory embeddings.

\item Empirical evaluation and ablation analysis of PGS on the design-bench benchmark. The code for PGS is publicly available at \url{https://github.com/yassineCh/PGS}.

\end{itemize}

\section*{Problem Setup}
\label{sec:problem-setup}

Suppose $\mathcal{X}$ is an input space where each $x \in \mathcal{X}$ is a $d$-dimensional candidate input. 
Let $f: \mathcal{X} \mapsto \Re$ be an unknown, expensive real-valued objective function which can evaluate any given input $x \in \mathcal{X}$ to produce output $y$ = $f(x)$. For example, in drug design application, $f(x)$ corresponds to running a physical lab experiment. Similarly, in hardware optimization application, $f(x)$ call involves performing an expensive 
simulation to mimic the real hardware. Our goal is to find an input $x \in \mathcal{X}$ that approximately optimizes $f$: 
\begin{equation}
\hat{x} \; \; \text{  s.t.  } \quad f(\hat{x}) \approx \max_x \; f(x)
\end{equation}
To solve this optimization problem, we are provided with a static dataset of $n$ input-output pairs $\mathcal{D}$=$\{(x_1, y_1), (x_2, y_2), \cdots,(x_n, y_n)\}$ collected {\em offline}, where $y_i$=$f(x_i)$. The optimization algorithm does not have access to objective function $f$ values on inputs outside the dataset $\mathcal{D}$. Hence, this problem is referred to as {\em offline black-box optimization}.

An offline model-based optimization algorithm produces an input $\hat{x} \in \mathcal{X}$ which is outside the training dataset $\mathcal{D}$. We measure the accuracy of solution in terms of the real objective function value of $\hat{x}$, namely $f(\hat{x})$. Ideally, $f(\hat{x})$ should be higher than the best function value seen in the offline dataset $\mathcal{D}$ (say $y_{best}$=$\max \{y_1, y_2,\cdots,y_n\}$). 

\section*{Related Work}
\label{sec:related} 

Prior work on offline optimization fall into two categories: 

\vspace{1.0ex}

\noindent {\bf Sampling from Generative Models.} Existing works in this family often focus on learning a generative model of the input space. For example, \cite{BrookeICML19} and \cite{ClaraNeurIPs20} employs a variational auto-encoder \cite{KingmaICLR14} which can be conditionally guided by a set of (domain-related) desirable properties. 
Similarly, \cite{MINS} learns an inverse mapping from the performance measure  to the input design using conditional generative adversarial network \cite{Mehdi14}.  
However, these approaches require learning a full generative model of the input space in addition to a surrogate model, which adds more complexity to the overall learning process. 
More recently, (BONET) \cite{krishnamoorthy2022generative} proposed a sequence modeling based conditional autoregressive model approach
that aims to mimic an online black-box optimizer represented by a collection of sorted trajectories synthesized from the offline data. 
However, BONET  requires knowledge of the oracle maxima 
which makes it unclear whether its performance remains robust if the assumed knowledge of the oracle maxima is not accurate. 

Both BONET and our proposed PGS rely on constructing synthetic trajectories from the given offline dataset. However, they differ in the methodology for trajectory synthesis. BONET uses sorted trajectories over the entire offline dataset while PGS generates randomized trajectories from top-$p$ percentile offline data. Unlike BONET, we provide a hyperparameter selection approach to tune $p$ for any given offline optimization task. Trajectories with monotonically increasing function values may not allow the policy to recover from mistakes, especially as the offline dataset is in a low function-value space. Trajectories with more variability in the function value as done in PGS
can overcome this challenge as we demonstrate through ablation studies.

\vspace{1.0ex}

\noindent {\bf Gradient-based Updates.}  
Gradient-based approaches advocate learning  a \emph{conservative} proxy that can be directly optimized via gradient ascent 
to circumvent the erroneous overestimation of the proxy model at out-of-distribution inputs. 
For example, \cite{YuNeurIPs21} mitigates the non-smooth nature of neural network using robust model pre-training and model adaptation to ensure a criteria of local smoothness while \cite{FuICLR21} uses normalized maximum likelihood to handle uncertainty in out-of-distribution (OOD) prediction. Alternatively, \cite{TrabuccoICML21} explicitly penalizes high-value prediction for OOD examples
to avoid erroneous overestimation of OOD input directly. Nonetheless, 
the optimization policies of these methods are mostly based on the local gradient information derived from the (imperfect) proxy at each data point, which in general does not always capture the (more global) relationship connecting the distance between two (arbitrary) input coordinates to the difference between their induced objective function values. This motivates us to investigate a reformulation of offline optimization as an offline reinforcement learning task which can naturally encode all the above information within a framework of (learnable) policy-guided gradient search, as detailed in Section~\ref{sec:method}.

\vspace{1.0ex}

\begin{figure*}[htp]
    \centering
    \includegraphics[width=0.7\textwidth]{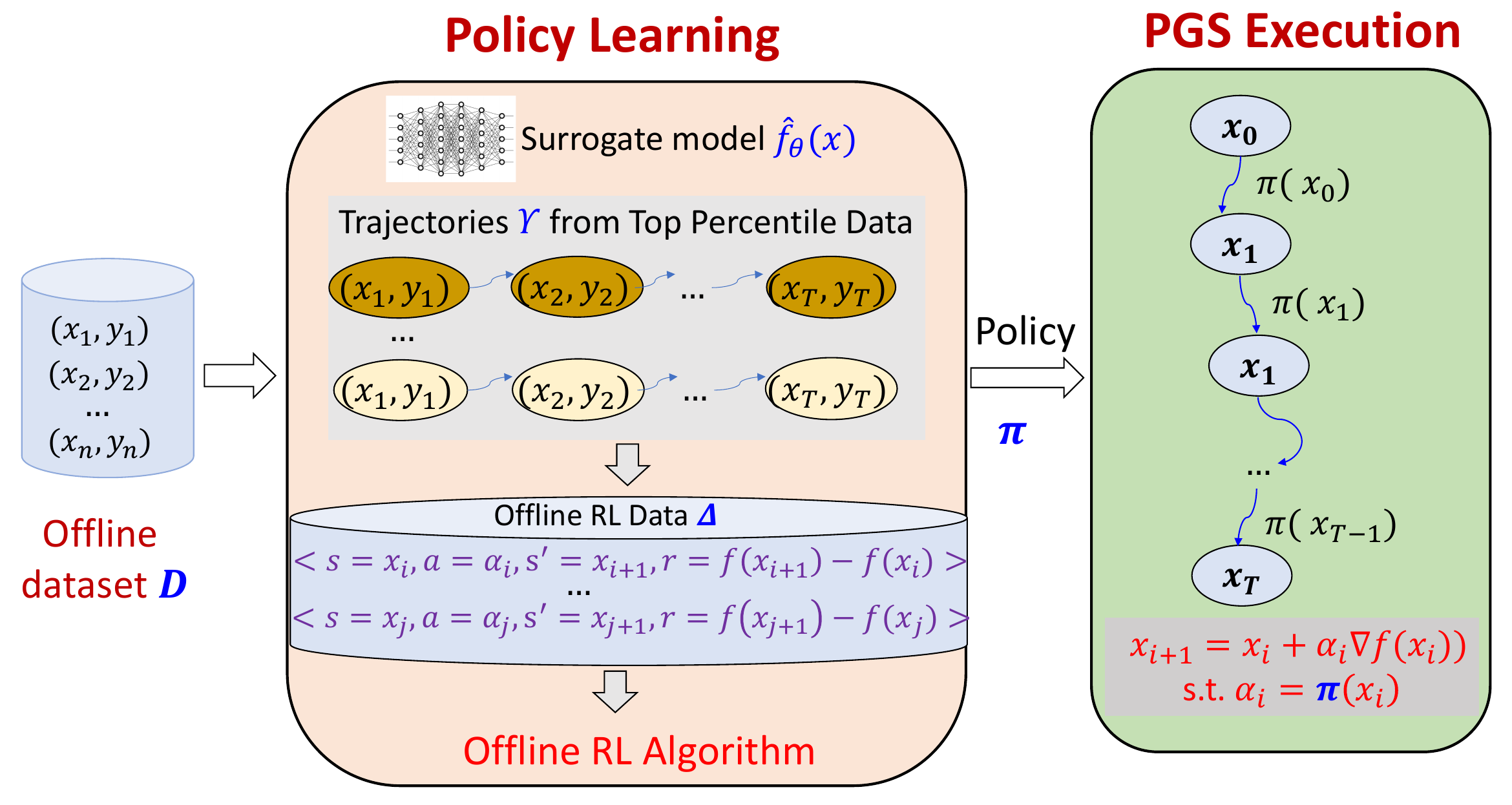}
    \caption{
        High-level overview of policy-guided gradient search approach for offline black-box optimization (BBO). The key idea is to cast the offline BBO problem as an offline RL problem. This reduction is accomplished by constructing random trajectories from a subset of inputs with high function values from the given offline data $\mathcal{D}$ (say Top $p$ percentile which is determined in a data-driven manner using our offline state estimation approach). The policy $\pi$ corresponds to selecting a step-size vector for a gradient based update on a trained surrogate model $\hat{f}_{\theta}$. Given a learned policy $\pi$, surrogate model $\hat{f}_{\theta}$ and a starting input $x_0$ with high function value sampled from the offline data $\mathcal{D}$, PGS performs $T$ steps of gradient search by asking the policy $\pi$ to predict the step-size vector $\alpha$ at each search step.}
    \label{fig:intro_fig}

\end{figure*}

\noindent {\bf Distinction from Learning-to-Optimize Setup:} Our policy learning problem is entirely different than the one considered in a seemingly related literature of learning-to-optimize (L2O) because we are only given a single offline dataset collected from a single task. Existing work on L2O \cite{andrychowicz2016learning} considers the problem setting where multiple datasets belonging to different tasks are available for the learner (aka meta-learning).

\section*{Policy-guided Gradient Search Algorithm}
\label{sec:method}

We first provide an overview of the PGS algorithm. Next, we describe a learning approach to create policies for PGS using offline RL. Finally, we outline the offline RL algorithm employed in our implementation and describe a methodology for hyper-parameter selection using only offline data.

\begin{algorithm}[t]
    \caption{Policy-guided Gradient Search (PGS)}
    \textbf{Input}: starting input for search $x_0 \in \mathcal{D}$; surrogate model $\hat{f}_\theta$; policy 
    $\pi$; number of search steps $T$ \\
    \textbf{Output}: best uncovered input $\hat{x}$
    \begin{algorithmic}[1]
        \FOR{each search step $k$=$1, 2,\cdots,T$}
        \STATE Predict step size: $\alpha_{k-1}$ = $\pi(x_{k-1})$
        \STATE Perform gradient step:  $x_{k} \leftarrow x_{k-1} + \alpha_{k-1} \nabla_x \hat{f}_\theta(x)|x=x_{k-1}$
        \ENDFOR
        \STATE \textbf{return} the solution of gradient search $x_T$
    \end{algorithmic}
    {\label{alg:PGS}}
\end{algorithm}

\noindent {\bf Overview of Policy-guided Gradient Search (PGS).} Our proposed offline model-based optimization approach PGS works as follows. PGS needs a surrogate model $\hat{f}_\theta: \mathcal{X} \mapsto \Re$ to make predictions on inputs outside the offline dataset $\mathcal{D}$ and a policy $\pi: \mathcal{X} \mapsto \Re{^d}$ to predict the step size to guide the gradient search towards inputs with high function values. PGS 
(Algorihtm \ref{alg:PGS}) performs $T$ steps of gradient search using both surogate model $\hat{f}_\theta$ and policy $\pi$ starting from an input $x_0 \in \mathcal{D}$ with high function value as follows: 
\begin{eqnarray}
\hspace{-6mm}x_{k} &\leftarrow& x_{k-1} \ +\ \alpha_{k-1} \nabla_x \hat{f}_\theta(x)|_{x=x_{k-1}} \ \text{with}\ k \in [T]
\end{eqnarray}
where $\alpha_{k-1} = \pi(x_{k-1})$ is the step-size predicted by policy $\pi$. The solution from the gradient search $x_T$ is returned as the output. The effectiveness of PGS for offline BBO critically depends on the policy $\pi$. We provide an offline RL formulation using the static dataset $\mathcal{D}$ to create robust policies to improve the accuracy of PGS. This reduction allows us to leverage the large body of work on offline RL to solve offline BBO problems in a principled manner.

\subsection*{Policy Learning Formulation}

We can train the surrogate function $\hat{f}_\theta$ from offline dataset $\mathcal{D}$ using a regression learner. The accuracy of PGS approach for solving offline BBO problem given a model $\hat{f}_\theta$, policy $\pi$, and starting input $x_0 \in \mathcal{D}$ is measured in terms of the real objective function value of $x_T$, i.e., $f(x_T)$. Therefore, the overall learning objective to create a policy to guide gradient search is as follows.
\begin{eqnarray}
\pi^* &\leftarrow& \max_{\pi \in \Pi} \mathbb{E}\left[f(x_T)\right] \ \ \text{such that}
\nonumber\\
x_0 &\sim& \mathcal{X}_{start} \ \ \text{and} \ \ x_T = \mathrm{PGS}\left(x_0, \hat{f}_\theta, \pi, T\right)
\end{eqnarray}

\noindent where $x_0$ is a starting input drawn from the distribution $\mathcal{X}_{start}$ and $x_T$ is the solution of PGS (Algorithm \ref{alg:PGS}) with 
surrogate model $\hat{f}_\theta$, policy $\pi$, and $T$ search steps. The optimal $\pi^*$ from the policy space $\Pi$ maximizes $f(x_T)$ in expectation (over $\mathcal{X}_{start}$). Before we discuss the details of our policy learning approach, we explain the corresponding Markov Decision Process (MDP) formulation. 

\vspace{1.0ex}

\noindent {\bf MDP Definition.} An MDP is a 4-tuple $<S, A, TF, R>$, where $S$ is the set of states, $A$ is the set of actions, $TF$ is a transition function, and $R$ is a bounded reward function. A policy $\pi: S \mapsto A$ is a mapping from states to actions. 

\vspace{0.5ex}

{\em State space $S$:} Every candidate input $x \in \mathcal{X}$ corresponds to one state $s \in S$.

\vspace{0.5ex}

{\em Action space $A$:} There are many choices for defining action space. Therefore, we first provide some motivation for the specific choice employed in our formulation. It is well-known in the optimization theory that second-order gradient based algorithms including Newton's method, Quasi-Newton's method are much more effective than first-order gradient methods in terms of reaching good solutions faster. The key idea in such methods is to employ second-order gradient (i.e., Hessian matrix) information to precondition the gradient in the solution iterate, i.e. $x_{k}$ = $x_{k-1} - B \nabla f(x_{k-1}))$ where $B$ is a preconditioning matrix. For example, $B$ is the inverse Hessian for Newton's method. However, as a consequence of requiring second-order gradients (Hessian), most of these approaches are computationally-expensive than first order gradient descent approaches. Inspired by this idea, we parameterize the action space of our gradient search in terms of a diagonal matrix of parameters $B$= diag$(\alpha)$ (i.e., a step-size vector) to strike a good balance between expressiveness and statistical efficiency for policy learning. Using a step-size vector (i.e., one parameter for each input dimension) is more expressive than the typical approach of using a scalar learning rate and more statistically efficient than an entire preconditioning matrix of parameters. This parametrization has also shown to be quite beneficial in the recent literature on learned optimizers or learning to learn methods \cite{flennerhag2019meta, li2016learning}. Hence, we consider each candidate action $a \in A$ to be a step-size vector $\alpha$. Note that each parameter value in $\alpha$ could be either positive or negative.

\vspace{0.5ex}

{\em Transition function $TF$:} The transition function $TF: S \times A \mapsto S$ is deterministic in our formulation. It takes a state $s$ = $x$ and an action $a$ = $\alpha$ to produce the next state $s'$ = $x'$. 
\begin{equation}
x' \leftarrow x + \alpha \nabla_x \hat{f}_\theta(x)
\end{equation}

\vspace{0.5ex}

{\em Reward function $R$:} The reward function $R: S \times A \mapsto \Re$ is defined as follows. The reward $R(s,a)$ of taking an action $a$ = $\alpha$ at a state $s$ = $x$ is equal to $f(x') - f(x)$, where $s'$ = $x'$ is the next state. In other words, reward is the difference between objective function values of next state and current state. It is positive when we go from states with low function values to states with high function values and vice versa. 

\vspace{1.0ex}

The above MDP construction is a deliberate design choice in our reduction approach to create policies using offline RL methods. In fact, improved offline optimization performance from PGS using this simple MDP in terms of state and reward representation demonstrates its effectiveness. We believe that considering more complex choices in the MDP (e.g., sequence modeling based unsupervised state embeddings and/or reward engineering) can be an interesting avenue for future research.

\subsection*{Trajectory Construction and RL Approach} 

Recall that we only have access to the function values for a subset of inputs from $\mathcal{X}$ as part of the offline dataset $\mathcal{D}$. We can employ $\mathcal{D}$ to create a set of random trajectories over the Top $p$ percentile data (sequence of inputs whose objective function values are high) and employ {{standard RL algorithms}} to create a policy $\pi$ for policy-guided search. We describe the two key steps of this approach below.

\vspace{1.0ex}

\noindent {\bf Trajectory synthesis.} We construct the set of trajectories from a prioritized subset $\mathcal{D}_{top}$ of inputs with high function values from the given offline dataset $\mathcal{D}$. This choice is primarily motivated by the need to align our training process with the test-time search algorithm. Recall that during the test-time, we perform $T$ gradient search steps guided by the trained policy starting from inputs with high objective function values. Naturally, it is useful to induce the policy in search regions from inputs with high function values as well. Therefore, we consider a simple and scalable choice of picking the examples lying in the $p^{th}$-percentile of the given offline dataset $\mathcal{D}$ sorted based on their objective function values. Note that our approach does not use any data outside the given offline dataset $\mathcal{D}$.

Restricting the trajectory synthesis to the subset $\mathcal{D}_{top}$ also allows more flexibility and robustness in constructing the trajectories for offline RL. We consider a simple approach to construct trajectories from $\mathcal{D}_{top}$. To create a trajectory of length $T$, $\tau$ = $(x_1, y_1), (x_2, y_2),\cdots, (x_T, y_T)$, we randomly sample a sequence of input-output pairs from $\mathcal{D}_{top}$ without replacement. We create a collection of trajectories $\Upsilon$=$\{\tau_1, \tau_2,\cdots,\tau_m\}$ by repeating this procedure several times. However, the decision to select the prioritized subset $\mathcal{D}_{top}$ has an inherent trade-off as this may lead to loss of diversity in the trajectories when we discard inputs from the low-ranked objective function values. Importantly, the right trade-off to achieve improved offline optimization performance may vary from one benchmark to another. Therefore, we provide a principled hyper-parameter selection methodology based on offline state estimation (explained later) to select the value of $p$ for Top $p$ percentile which can be potentially useful for other offline optimization approaches.

\vspace{1.0ex}

\noindent {\bf Baseline RL approach.} Given the set of trajectories $\Upsilon$=$\{\tau_1, \tau_2,\cdots,\tau_m\}$, a straightforward approach would be to use a standard RL algorithm  \cite{sutton2018reinforcement}
with a function approximator \cite{mnih2015human} 
to find a policy $\pi$.  Although this approach is very simple, it will most likely fail as the learned policy $\pi$ cannot handle out-of-distribution (OOD) states/inputs, i.e., inputs outside the offline dataset $\mathcal{D}$.  Instead, we propose using an offline RL approach which leverages explicit conservatism to find policies which are robust to OOD data. Indeed, our experiments show that incorporating OOD robustness via offline RL is critical for good performance on offline BBO.

 \begin{algorithm}[t]
    \caption{Learning to Guide Gradient Search}
    \footnotesize
   \textbf{Input}: offline dataset $\mathcal{D}=\{(x_i,y_i)\}_{i=1}^{n}$; number of input dimensions $d$; number of search steps $T$ \\
    \textbf{Output}: model $\hat{f}_\theta$ and policy $\pi$
    \begin{algorithmic}[1]
        \STATE Learn surrogate model $\hat{f}_\theta$: $\hat{f}_\theta \leftarrow $ \textsc{Regression-Learner}($\mathcal{D}$)
        \STATE Select $\mathcal{D}_{top} \subseteq \mathcal{D}$ consisting of inputs with Top $p$ percentile highest objective values 
        \STATE Create a collection of random trajectories of length $T$ using $\mathcal{D}_{top}$:  $\Upsilon$=$\{\tau_1, \tau_2,\cdots,\tau_m\}$ s.t every trajectory $\tau_i$ consists of a random sequence of examples from the Top $p$ percentile data 
        \STATE Training data for policy learning $\Delta \leftarrow \emptyset$
        \FOR{each trajectory $\tau$=$(x_1,y_1),\cdots,(x_T,y_T)) \in \Upsilon$}
        \FOR{each pair $(x_{k-1},y_{k-1})$ and $(x_{k},y_{k})$ from $\tau$}
                 \STATE Add a sample $(s, a, s', r)$ to $\Delta$ \\ where current state $s$ = $x_{k-1}$, next state $s'$ = $x_{k}$, reward $r$ = $y_{k}-y_{k-1}$, and action $a$ = $\alpha_{k-1}$ s.t $x_{k} \leftarrow x_{k-1} + \alpha_{k-1} \nabla_x \hat{f}_\theta(x)|x=x_{k-1}$
        \ENDFOR                
        \ENDFOR
        \STATE Learn policy $\pi$: \\ $\pi \leftarrow $ \textsc{Offline-Reinforcement-Learner}($\Delta$)
         \STATE \textbf{return} the learned surrogate function $\hat{f}_\theta$ and policy $\pi$
    \end{algorithmic}
    {\label{alg:training}}
\end{algorithm}

\subsection*{Offline RL Approach for Policy Learning} 

Prior work on offline RL has shown significant successes in learning policies from a given offline dataset of experiences (i.e., 4-tuples consisting of state, action, next state, and reward). Inspired by these successes, we consider an offline RL approach to learn policies that are robust to OOD inputs outside the offline data $\mathcal{D}$. In our MDP formulation, we do not know the reward for some pairs of states, when we do not know the real objective function value of the corresponding input of at least one of the states.

To learn an effective policy to handle OOD states via offline RL, an agent should deviate from the available behavior in the logged/offline data. However, distributional shift scenarios are overestimated by the value function, leading to mediocre policies. Model-free approaches circumvent such bias by embedding constraints over the objective function via some form of regularization. 
Constraints include the policy, the value function, or both \cite{fujimoto2019benchmarking, peng, kumar19, wu2020behavior, CQL, FujimotoG21}. Model-based approaches create conservative MDPs of the environment and proceed to apply regular online RL methods since they have access to the environment dynamics \cite{Morel, MOPO, Combo}.

Our offline RL approach (see Algorithm \ref{alg:training}) shares the first step of the above-mentioned baseline RL approach, namely, trajectory synthesis. We employ the dataset of random trajectories $\Upsilon$=$\{\tau_1, \tau_2,\cdots,\tau_m\}$ to create a training set which is appropriate for offline RL as explained below. 

\vspace{1.0ex}

\noindent {\bf Reduction to offline RL.} Recall that each training example for offline RL is a 4-tuple of the form $(s, a, s', r)$, where $s$ is a current state, $a$ is an action, $s'$ is the next state resulting from taking an action $a$ at state $s$, $r$ is the reward for going to next state $s'$ from $s$. We generate a set of training examples of this form using random trajectories as follows (Line 5-9 in Algorithm \ref{alg:training}). For each trajectory $\tau \in \Upsilon$ = $(x_1, y_1), (x_2, y_2),\cdots, (x_T, y_T)$, we create one offline RL training example for every pair of inputs $(x_{k-1}, y_{k-1})$ and $(x_k, y_k)$ from $\tau$. The state $s$ = $x_{k-1}$, the action $a$ = $\alpha$ such that the gradient step will result in the next state $s'$ = $x_k$: $x_k \leftarrow x_{k-1} + \alpha \nabla_x \hat{f}_\theta(x)|x=x_{k-1}$, the next state $s'$ = $x_k$, and the reward $r$ = $f(x_k)-f(x_{k-1})$. The aggregate set of offline RL training examples $\Delta$ over all trajectories in $\Upsilon$ are given to an offline RL algorithm to create the policy $\pi$ (Line 10 in Algorithm \ref{alg:training}). The main advantage of this reduction is that it allows us to leverage prior work on offline RL to solve the challenging problem of offline black-box optimization.

\vspace{1.0ex}

\noindent {\bf Conservative Q-learning (CQL).} To implement our PGS approach, we can employ any existing offline RL algorithm. We chose conservative Q-learning because of its demonstrated effectiveness in practice \cite{CQL}. For the sake of completeness, we briefly describe the CQL approach and how it is configured for our experimental evaluation. To address distributional shift, CQL learns a lower bound of the true policy value leading to a more reliable value function estimate. Along with the standard temporal difference error, a regularizer is added to minimize Q-values for unseen actions sampled from a distribution $\mu$ -- see Eq.~\eqref{cql}, which optimizes for $\hat{Q}^\pi = \arg\min_Q \ell(Q)$ where

\begin{eqnarray}
\label{cql}
\hspace{-1mm}\ell(Q) \hspace{-2mm}&\triangleq& \hspace{-2mm}\max_{\mu(a|s)}\left(\mathop{\mathbb{E}}_{s\sim\mathcal{D}}\mathop{\mathbb{E}}_{a\sim\mu}\Big[Q(s,a)\Big] -\hspace{-2mm} \mathop{\mathbb{E}}_{(s,a)\sim\mathcal{D}}\Big[Q(s,a)\Big]\right)\nonumber\\
\hspace{-2mm}&+& \frac{1}{2\beta}\mathop{\mathbb{E}}_{(s,a,s') \sim \mathcal{D}}\Big[\Delta(s,a,s')\Big]
\end{eqnarray}
with $\Delta(s, a, s') = (r(s, a) + \gamma\mathbb{E}_\pi[Q(s',a')] - Q(s, a))^2$

The conservative Q function is then used to train a soft-actor critic agent in our case \cite{SAC}. Since our MDP is deterministic in nature, CQL's theoretical analysis is also directly applicable to the offline optimization setting. This is another advantage of our reduction formulation!

\subsection*{ Offline State Estimation via Latent Embeddings for Hyperparameter Selection} 

We propose a principled methodology for selecting the key hyper-parameters of our PGS approach. The main idea is to assign estimated function values to unknown inputs (outside the offline dataset) such that they are mostly correlated with the true function values. We leverage offline state representation learning objectives \cite{yang2021representation, oh2017value} to embed the inputs (states in our MDP formulation) into a latent embedding space. This idea of pre-trained state representations from offline data has been shown to be quite effective for RL algorithms \cite{yang2021representation}. To get an offline estimate of the performance of inputs suggested by PGS, we employ the following algorithmic procedure which we refer to as {\em Offline State Estimation via Latent embeddings (OSEL)}: 
\begin{itemize}
    \item Embed unknown input into the latent embedding space.
    \item Get a K-nearest neighbor regressor based estimate of the performance, where we pick $K$ nearest offline dataset inputs in the latent space as neighbors.
\end{itemize}

We would like to emphasize that the OSEL procedure only uses the offline dataset $\mathcal{D}$ and does not rely on external sources. We perform pre-training of embeddings using random trajectories generated from the entire offline dataset. Our hyper-parameter selection methodology represents a thoughtful and principled means of improving the quality of input representations, which, in turn, contributes to the overall performance of our approach.

\section*{Experiments and Results}
\label{sec:experiments}

We first describe our setup including the benchmarks, evaluation methodology, and baselines. Next, we discuss the results comparing the PGS approach with baseline methods.

 \begin{table*}[ht!] 
    \centering
    \begin{tabular}{ccccc}
		\specialrule{\cmidrulewidth}{0pt}{0pt}
		Continuous Tasks  & Superconductor & Ant Morphology & D'Kitty Morphology \\
		\specialrule{\cmidrulewidth}{0pt}{0pt}
		\textbf{PGS w/ Offline RL} & \textbf{0.563} $\pm$ \textbf{0.058}  &  \textbf{0.949} $\pm$ \textbf{0.017} &  \textbf{0.966} $\pm$ \textbf{0.013} \\
		PGS w/ Standard RL & 0.528 $\pm$ 0.012 & 0.303 $\pm$ 0.017 & 0.900 $\pm$ 0.007\\
	\end{tabular}
 
	\begin{tabular}{ccccc}
		\specialrule{\cmidrulewidth}{0pt}{0pt}
		Discrete Tasks & GFP & TF Bind 8 & UTR  \\
		\specialrule{\cmidrulewidth}{0pt}{0pt}
		\textbf{PGS w/ Offline RL} & \textbf{0.864} $\pm$ \textbf{0.000} & \textbf{0.981} $\pm$ \textbf{0.015} & \textbf{0.713} $\pm$ \textbf{0.009}  \\
		PGS w/ Standard RL & \textbf{0.864} $\pm$ \textbf{0.000} & 0.774 $\pm$ 0.063 & 0.686 $\pm$ 0.013 \\
	\end{tabular}
 \caption{Ablation Results showing the ablation of using standard RL as opposed to offline RL to create the policy for PGS.} \label{tab: conservatism}
 
\end{table*}

\subsection*{Experimental Setup}

\noindent {\bf Benchmark tasks.} We employ six challenging benchmark tasks (and corresponding datasets) from diverse domains.  
All these datasets and the oracle evaluations are accessed via the design-bench benchmark \cite{TrabuccoArXiv22}. {\em \bf  1) D'Kitty Morphology:} This task requires optimizing the morphology of a four-legged robot named D'Kitty in order to reach a given location \cite{ahn2020robel}. The morphological parameters of the robot is defined over a {\em 56-dimensional continuous} search space. {\em \bf   2) Ant Morphology:} Similar to the D'Kitty benchmark, this task \cite{brockman2016openai} requires optimizing the morphology of a 3D robot.  
 The search space of parameters is continuous and 60-dimensional in size. {\em \bf   3) Superconductor:} This task \cite{BrookeICML19} involves designing superconductors with high critical temperatures which has many engineering  and material science applications and each input point is represented by a 86-dimensional continuous vector. {\em  \bf  4) GFP:} The goal of this task \cite{rao2019evaluating} is to find fluorescence maximizing proteins. Each point in the search space denotes a protein which is represented by a 237-dimensional vector. Each input dimension can take values from 20 different amino acids. {\em \bf 5) TF Bind8:} This task aims to maximize the binding activity score between a given human transcription factor and a DNA sequence. It's a discrete task with 8-length DNA sequences represented as vectors with values from 4 categories.
 {\em \bf 6) UTR:} This task \cite{sample2019human} requires optimizing a search space of 50-dimensional DNA sequences to maximize the expression level of a given gene.

\begin{table*}[ht!] 
 \centering
	
	\scalebox{1.}{
	\begin{tabular}{lccc}
		\specialrule{\cmidrulewidth}{0pt}{0pt}
		Method & Superconductor & Ant Morphology & D’Kitty Morphology\\
		\specialrule{\cmidrulewidth}{0pt}{0pt}
		$\mathcal{D}$(\textbf{best}) & $0.399$ & $0.565$ & $0.884$  \\
		BO-qEI & 0.402 $\pm$ 0.034 & 0.819 $\pm$ 0.000 & 0.896 $\pm$ 0.000 \\
		CMA-ES & 0.465 $\pm$ 0.024 & \textbf{1.214 $\pm$ 0.732} & 0.724 $\pm$ 0.001 \\
		REINFORCE & 0.481 $\pm$ 0.013 & 0.266 $\pm$ 0.032 & 0.562 $\pm$ 0.196 \\
		CbAS & 0.503 $\pm$ 0.069 & 0.876 $\pm$ 0.031 & 0.892 $\pm$ 0.008 \\
		Auto.CbAS & 0.421 $\pm$ 0.045 & 0.882 $\pm$ 0.045 & 0.906 $\pm$ 0.006 \\
		MIN & 0.469 $\pm$ 0.023 & 0.913 $\pm$ 0.036 & 0.945 $\pm$ 0.012 \\
        BONET & $0.411 \pm 0.024 $ & 0.927 $\pm$ 0.010 & 0.954 $\pm$ 0.009\\
		\specialrule{\cmidrulewidth}{0pt}{0pt}
		Grad & 0.518 $\pm$ 0.024 & 0.293 $\pm$ 0.023 & 0.874 $\pm$ 0.022 \\
		COMs & 0.439 $\pm$ 0.033 & 0.944 $\pm$ 0.016 & 0.949 $\pm$ 0.015 \\
		ROMA & $0.476 \pm 0.024$ & $0.814 \pm 0.051$ & $0.905 \pm 0.018$ \\
        NEMO & ${0.488} \pm {0.034}$ & $0.814 \pm 0.043$ & $0.924 \pm 0.012$ \\
        BDI & $0.520 \pm 0.005 $ & $0.962 \pm 0.000$ & $0.941 \pm 0.000$ \\
		\specialrule{\cmidrulewidth}{0pt}{0pt}
		\specialrule{\cmidrulewidth}{0pt}{0pt}
		\textbf{PGS (ours)}   & $\textbf{0.563} \pm \textbf{0.058} $ & $ 0.949 \pm 0.017$ & $\textbf{0.966} \pm \textbf{0.013}$ \\
		\specialrule{\cmidrulewidth}{0pt}{0pt}
	\end{tabular}
	}
 \caption{Results comparing PGS and baseline methods on benchmark tasks with continuous search spaces. PGS finds better solutions than all other approaches on Superconductor and D'Kitty Morphology and is competitive on Ant Morphology.} \vspace{0.1cm} \label{tab: continuous}
\end{table*} 
\begin{table*}[ht!]  
	\centering
	\begin{tabular}{lcccc}
		Method & GFP & TF Bind 8 & UTR & \textbf{Mean Rank} \\
		\specialrule{\cmidrulewidth}{0pt}{0pt}
		$\mathcal{D}$(\textbf{best}) & $0.789$ & $0.439$ & $0.593$  \\
		BO-qEI & 0.254 $\pm$ 0.352 & 0.798 $\pm$ 0.083 & 0.684 $\pm$ 0.000 & $11.33/13$\\
		CMA-ES & 0.054 $\pm$ 0.002 & 0.953 $\pm$ 0.022 & 0.707 $\pm$ 0.014 & $7/13$\\
		REINFORCE & \textbf{0.865 $\pm$ 0.000} & 0.948 $\pm$ 0.028 &          0.688 $\pm$ 0.010 & $8.16/13$\\
		CbAS & \textbf{0.865 $\pm$ 0.000} & 0.927 $\pm$ 0.051 & 0.694 $\pm$ 0.010 & $6.33/13$\\
		Auto.CbAS & \textbf{0.865 $\pm$ 0.000} & 0.910 $\pm$ 0.044 & 0.691 $\pm$ 0.012 & $7.5/13$\\
		MIN & \textbf{0.865 $\pm$ 0.001} & 0.905 $\pm$ 0.052 & 0.697 $\pm$ 0.010 & $5.83/13$ \\
        BONET & $\textbf{0.864 $\pm$ 0.000}$ & 0.911 $\pm$ 0.034 & 0.688 $\pm$ 0.011 & $7.33/13$ \\
		\specialrule{\cmidrulewidth}{0pt}{0pt}
		Grad & \textbf{0.864 $\pm$ 0.001} & 0.977 $\pm$ 0.025 & 0.695 $\pm$ 0.013 & $6.5/13$ \\
		COMs & \textbf{0.864 $\pm$ 0.000} & 0.945 $\pm$ 0.033 & 0.699 $\pm$ 0.011 & $5.33/13$\\
		ROMA & $0.558 \pm 0.395$ & $0.928 \pm 0.038$ & $0.690 \pm 0.012$ & $8.66/13$ \\
        NEMO & 0.150 $\pm$ 0.270 & 0.905 $\pm$ 0.048 & 0.694 $\pm$ 0.015  & $8.5/13$ \\
        BDI & \textbf{0.864 $\pm$ 0.000} & 0.973 $\pm$ 0.000 & \textbf{0.760 $\pm$ 0.000} & 3/13 \\
		\specialrule{\cmidrulewidth}{0pt}{0pt}
		\textbf{PGS (ours)}  & \textbf{0.864 $\pm$ 0.000} & \textbf{0.981 $\pm$ 0.015}  & 0.713 $\pm$ 0.009  & \textbf{2.16/13}\\
		\specialrule{\cmidrulewidth}{0pt}{0pt}
	\end{tabular}
 \caption{Results comparing PGS and baselines on benchmarks with discrete input spaces. PGS finds better solution than all other approaches on TF Bind 8, reaches the best benchmarks solution on GFP and is competitive on UTR.  The last column show the mean rank computed across all the 6 tasks to measure the overall effectiveness of methods for multiple tasks.}\vspace{0.2cm}  \label{tab: discrete}
\end{table*}

\vspace{1.0ex}

\noindent {\bf Configuration of algorithms and baselines.}
We compare PGS against baselines including COMs \cite{TrabuccoICML21}, NEMO \cite{FuICLR21}, ROMA \cite{YuNeurIPs21}, BDI \cite{BDI}, BONET \cite{BONET} and other baselines from design-bench \cite{TrabuccoArXiv22}.
We took the results for all baselines from their respective papers. Since NEMO and ROMA don't report normalized scores, and their original papers lack a few tasks, we take their results from the state-of-the-art BDI paper \cite{BDI}. This is reasonable because all baselines use the same design-bench benchmark and evaluation methodology. We note that 
BONET utilizes double the evaluation budget (i.e., 256 points) compared to all other methods. Therefore, for fair evaluation with all the baselines, we ran BONET code (from official implementation {\url{https://github.com/siddarthk97/bonet}}) to generate 128 points for evaluation.

We configure PGS as follows. For each task, we normalize inputs and outputs before we train a vanilla multilayer perceptron, $\hat{f}_\theta$,  with two hidden layers with 2048 units and ReLU activation. $\hat{f}_\theta$ is trained to minimize the mean squared error of function values. 
We employ the publicly available implementation of CQL \url{https://github.com/young-geng/CQL}. PGS is based on gradient updates over continuous valued inputs in contrast to discrete tasks. To mitigate this, we learn a latent representaion of the discrete inputs. We apply a variational autencoder and train PGS over the learned latent space. Note that we decode the results before oracle evaluation.

We configure OSEL for hyper-parameter selection as follows. We employ contrastive predictive loss objective as used in value prediction networks (VPNs) \cite{oh2017value} where the key idea involves learning to predict $m$-step future rewards and value functions starting from a given state. We used the publicly available implementation of VPNs (\url{https://github.com/google-research/google-research/tree/master/rl_repr}).
We employ 20000 trajectories of length $T$=50 to align training with test-time search process. We evaluate four different values of $p$ = $\{10, 20, 30, 40\}$ for Top $p$ percentile data and number of epochs of CQL ranging from 50 to 400 in increments of 50 and picked the configuration with the best OSEL performance. 

\vspace{1.0ex}

\noindent {\bf Evaluation methodology.} We follow the methodology as introduced in the design-bench benchmark \cite{TrabuccoArXiv22} and employed by all prior work on offline BBO. Each algorithm generates a set of $N$ points which are evaluated by oracle and the 100th percentile (best value among the $N$ points) is computed to compare the different approaches. All objective/oracle values are normalized to $[0, 1]$ by using the mean and max from a larger unobserved data set. We run PGS and BONET on each task for five different runs and report the mean and standard deviation in the results section.

\subsection*{Results and Discussion}
\noindent {\bf PGS w/ offline RL vs. standard RL.}
To clearly show that the offline RL component of PGS approach brings significant advantage, we perform ablation by replacing it with a standard RL algorithm while keeping everything else same including the synthesized trajectories. We chose soft-actor critic as the RL method. This allows us to keep it consistent with our offline RL method (CQL) since it builds upon soft-actor critic as one of it's components. Results in Table \ref{tab: conservatism} show that the performance of PGS becomes consistently worse by using standard RL over CQL, which demonstrates the importance of employing offline RL algorithm to correct for OOD data using the principle of conservatism.
\vspace{1.0ex}

\noindent {\bf Comparison with state-of-the-art.} We present the results comparing PGS with all  baselines in Tables \ref{tab: continuous} (continuous input spaces) and \ref{tab: discrete} (discrete input spaces).  
Each column with the task name shows the 100th percentile (top score) found by each method on the corresponding task. The $\mathcal{D}$(\textbf{best}) row refers to the highest score in each task's offline data.  We also show the mean rank achieved by the methods computed across all the tasks.  We highlight in bold the methods with the best results for every task.

\begin{table*}[ht!]
  \centering
  \begin{tabular}{lllll}
    \toprule
    Task  & AntMorphology  & DKittyMorphology & Superconductor & TFBind8  \\
    \midrule
    {PGS (IQL)} & ${0.939} \pm {0.021}$ & ${0.959} \pm {0.002}$ & $ 0.532  \pm 0.000$ &  0.960 $\pm$ 0.027 \\
    PGS (CQL) & $ 0.949 \pm 0.017$ & ${0.966} \pm {0.013}$ & ${0.563} \pm {0.058} $ & 0.981 $\pm$ 0.015 \\
    \bottomrule
  \end{tabular}
  \caption{Table comparing the results of training with another offline RL algorithm: IQL \cite{KostrikovNL22}}  \label{tab:main_cql_vs_iql}

\end{table*}

\begin{table*}[ht!]
  \centering
  \begin{tabular}{lllll}
    \toprule
    Task  & AntMorphology  & DKittyMorphology & Superconductor & TFBind8  \\
    \midrule
    PGS (entire data) & $0.890 \pm 0.022$ & $0.947 \pm 0.013$ & $0.533 \pm 0.036$ & $0.955 \pm 0.024$ \\
{PGS (top $p$)}    & \textbf{0.949 $\pm$ 0.017} & \textbf{0.966 $\pm$ 0.013} & $ \textbf{0.563 $\pm$ 0.058} $ & \textbf{0.981 $\pm$ 0.015} \\
    \bottomrule
  \end{tabular}
  \caption{Table comparing the results of PGS using trajectories synthesized with top $p$ percentile data vs. entire offline dataset.}
  \label{tab:main_top_p_vsentire}
\end{table*} 

\begin{table*}[ht!]
  \centering
  \begin{tabular}{llll}
    \toprule
    Task & AntMorphology  & DKittyMorphology & TFBind8  \\
    \midrule
    PGS (monotonic) & $0.916 \pm 0.027$ & $0.964 \pm 0.013$ & $0.925 \pm 0.031$ \\
    PGS (top p) & \textbf{0.949 $\pm$ 0.017} & \textbf{0.966 $\pm$ 0.013} & \textbf{0.981 $\pm$ 0.015} \\
    \bottomrule
  \end{tabular}
\caption{Table comparing the results of PGS using top p random trajectories vs. monotonically increasing trajectories}
  \label{tab:main_PGS_vs_BONET}
\end{table*}

Cumulatively, PGS achieves an average ranking of 2.16 across all tasks, which is higher than all the baselines. Individually, PGS finds the best scores on 4 (Superconductor, D'Kitty Morphology, TFBind8, and GFP) out of the 6 tasks and very close to the best baseline on Ant Morphology and UTR. PGS significantly outperforms baselines, especially in Superconductor and D'Kitty tasks, and excels in surpassing the best designs from offline data across all tasks, showcasing its effectiveness in offline BBO problem-solving. 

Since the performance of offline RL critically depends on the amount of state space coverage in the offline dataset, one potential limitation of PGS is that it may not perform well if the coverage is bad. For example, the offline dataset for the Ant morphology task are picked from the lowest scoring parts of the objective space relative to other benchmarks.

\vspace{1.0ex}

\noindent {\bf Ablation experiments.} We conducted several ablations to demonstrate the effectiveness and generality of PGS. 

\begin{itemize}
    
    \item {\em PGS with other offline RL methods:} We conducted ablation experiments on PGS by replacing CQL with another off-the-shelf offline RL algorithm, implicit Q learning (IQL) \cite{KostrikovNL22}. Remarkably, we observed consistent and strong performance across various tasks. These results in Table \ref{tab:main_cql_vs_iql}, averaged over five runs, underscore the robustness of PGS's core concept: the reduction to an offline RL problem. 
    
    \item {\em PGS with Top-$p$ percentile vs. entire offline data:} We run the ablation of running offline RL on trajectories constructed from top $p$ prioritized data subset versus that from the entire offline dataset. The results shown in Table \ref{tab:main_top_p_vsentire} empirically confirm that the top $p$ percentile subset selection is a better strategy for trajectory construction.

    \item {\em PGS with monotonic trajectories over entire offline data:} We perform PGS ablation to compare our proposed trajectory synthesis approach (random trajectories over the top $p$ percentile offline data) with BONET (monotonic trajectories over the entire offline data). The results shown in Table \ref{tab:main_PGS_vs_BONET} provide empirical evidence that our trajectory design approach reaches better solutions. 

    \item {\em Other ablations:} We also performed ablation experiments on variations in number of gradient search steps at test-time and examining the performance of PGS with varying dataset sizes for offline RL. Furthermore, 
    We analyzed the policy-guided search trajectory at test-time by computing the norms of action vectors. We observed that the norms of action sequence consistently decrease as the number of search steps increase (aggressive to conservative exploration of search space). This phenomenon suggests that our learned policy becomes more adept at guiding the search as we approach regions associated with high-value designs. For more details on all  ablation experiments, please refer to the Appendix.

\end{itemize}

\section*{Summary and Future Work}
\label{sec:conclude}

This paper introduces the new perspective of policy-guided gradient search for offline black-box optimization (BBO). This perspective is aimed at improving the search strategy in offline BBO, which complements prior methods that have focused on improving surrogate models
while using fixed search strategies. Empirical results show that learned search strategies can help in improving the accuracy of optimization significantly across multiple benchmarks. There are many avenues of future work. First, exploring more sophisticated trajectory sampling methods will improve the effectiveness of many offline optimization methods including PGS and BONET. Second, while our hyperparameter selection method based on offline state estimation showed promise, but finding the appropriate hyper-parameters for offline optimization remains an open and important research challenge. Finally, identifying suitable offline optimization methods for specific problems based on problem properties is a valuable future research direction which can benefit practitioners and motivate researchers to drive algorithmic advancements to improve the Pareto front of solutions.

\section*{Acknowledgements} The authors gratefully acknowledge the in part support from National Science Foundation (NSF) grants IIS-1845922 and CNS-2308530. The views expressed are those of the authors and do not reflect the official policy or position of the NSF.

\bibliography{aaai24}
\clearpage

\appendix
\section{Appendix}
In this section, we provide more details about the experimental setup and include ablation results to further evaluate our PGS approach. The hyper-parameters for learning the surrogate model and OSEL embeddings are given in Table \ref{tab:app_sm_hp}. They are kept same as the previous related work and are fairly robust.

\begin{table*}[h!]
    \centering
    \caption{Hyperparameters used for training the surrogate model and latent embedding for the OSEL procedure.}
\label{tab:app_sm_hp}
    \begin{tabular}{lcc}
        \toprule
        Hyperparameter & Discrete & Continuous \\
        \midrule
        Number of epochs to train $\hat{f}_\theta$ & 50 & 50 \\
        Batch size to train $\hat{f}_\theta$ & 128 & 128 \\
        Adam learning rate to train $\hat{f}_\theta$ & $3\mathrm{e}{-4}$ & $3\mathrm{e}{-4}$ \\
        $T$ Number of gradient ascent steps in Equation 3 & 50 & 50 \\
        Discrete tasks latent vector dimension & 32 & - \\
        Scale of $\alpha$ in Equation 4  & $2.0 \sqrt{d}$ & $0.05 \sqrt{d}$\\
        KNN regressor k & 10 & 10 \\
        KNN regressor k (breaking ties) & 100 & 100 \\
        OSEL latent space dimension  & 8 & 32 \\
        OSEL embedding training window  & 8 & 8 \\
        \bottomrule
    \end{tabular}
\end{table*}
\subsection{Hyper-parameter selection via OSEL}
\label{gen_inst}
In this section, we describe the step-by-step procedure involved in selecting two key hyperparmeters of PGS: top $p$ percentile of the data for trajectory construction and number of epochs required for training the offline RL policy. Here, the number of epochs are conditionally dependent on the choice of $p$ hyper-parameter. For each candidate $p$ value in $\{10, 20, 30, 40\}$, we execute following steps: i) Generate a set of trajectories, ii) Train the offline RL (CQL) algorithm for 400 epochs while logging the agent at every 50 epochs resulting in a total of 8 policies, iii) Perform multiple (128) policy guided searches with each candidate logged agent which is exactly similar to the test-time evaluation, iv) Apply OSEL to evaluate the final designs suggested by policy guided search for each agent and compute their average as the final score.  This procedure allows us to compute the score for each element in the cartesian product of $p = \{10, 20, 30, 40\}$ and no of epochs $\{50, 100, 150, 200, 250, 300, 350, 400\}$. We pick the hyper-parameter pair with the best average OSEL score. The values chosen by our hyper-parameter selection procedure for each task are reported in Tables \ref{tab:pgs_osel_hp} and \ref{tab:iql_osel_hp}.

 \begin{table*}[h!]
  \caption{Selection of top $p$ percentile and number of epochs for CQL training to train the policy for PGS for each task employed in the experiments.}
  \label{tab:pgs_osel_hp}
  \centering
  \begin{tabular}{lllllll}
    \toprule
    Task  & Ant & DKitty & Superconductor & TFBind8  & UTR & GFP\\
    \midrule
    Top p & $20$ & $40$ & $40$ & $30$ & $10$  & $40$   \\
    CQL training epochs & $300$ & $300$ & $100$ & $150$ & $50$ & $50$ \\
    \bottomrule
  \end{tabular}
\end{table*}

\subsection{Discrete Tasks VAE details}
We employ a $\beta$-VAE \cite{higgins2016beta} framework with $\beta$ = 1.0 to train an encoder and a decoder network. Both the encoder and decoder networks consist of four residual convolution blocks. Each block comprises two 1D convolution layers with a hidden size of 64 and a kernel size of 3. We use the publicly available code of $\beta$-VAE, from the design baselines, and train for 50 epochs.

\subsection{Ablation experiments and results}
As mentioned in the main text, we present additional ablation results here. 

\subsubsection{PGS test-time action norms} 
\begin{figure}[htp]
    \centering    \includegraphics[width=0.47\textwidth]{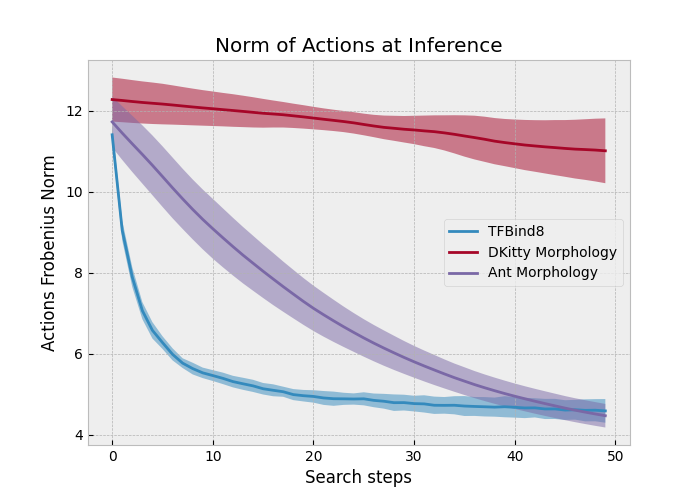}
    \caption{PGS Action Norms During Search Steps}
    \label{fig:norms_fig}

\end{figure}

We examined the norms of actions generated by the offline RL algorithm, as depicted in figure \ref{fig:norms_fig}, per the number of search steps. Clearly, the action norms decrease as the search progresses. We hypothesize that this trend arises from the offline RL algorithm's improved guidance of the search process, encouraging more conservative actions as we approach regions near the optimal design.

\begin{table*}[h!]
  \caption{Selection of top $p$ percentile and number of epochs for IQL training to train the policy for PGS for each task employed in the experiments.}
  \label{tab:iql_osel_hp}
  \centering
  \begin{tabular}{lllll}
    \toprule
    Task  & Ant & DKitty & Superconductor & TFBind8 \\
    \midrule
    Top p & $40$ & $20$ & $40$ & $40$   \\
    IQL training epochs & $100$ & $50$ & $100$ & $300$  \\
    \bottomrule
  \end{tabular}
\end{table*}

\subsubsection{Test-time trajectory length (number of gradient search steps) } We ablate the number of search steps policy guided search is run during test-time with a policy trained for 50 search steps and show the results in Table \ref{tab:app_eval_steps}. Recall that we employ this search starting from 128 different points generating a set of 128 points for evaluation. Based on the results, we see that the performance drops as we increase the no. of steps (trajectory length). This decrease in performance can be attributed to the train-test mismatch in offline RL policy training since our policies are trained for trajectories of fixed length 50, which is aligned with the standard evaluation procedure in the offline optimization literature.

\begin{table*}
  \caption{Table showing the ablation result of comparing different trajectory lengths (number of search steps at the test-time) guided by policy guided search during test-time search.}
  \label{tab:app_eval_steps}
  \centering
  \begin{tabular}{lllllll}
    \toprule
    \# Gradient Steps  & 50  & 60 & 70 & 80 & 90 & 100 \\
    \midrule
    AntMorphology & $0.95 \pm 0.02$ & $0.91 \pm 0.02$ & $0.91 \pm 0.02$ & $0.91 \pm 0.03$  & $0.92 \pm 0.04$  & $0.89 \pm 0.04$  \\
    DKittyMorphology & $0.97 \pm 0.01$ & $0.96 \pm 0.01$ & $0.96 \pm 0.02$ & $0.96 \pm 0.01$  & $0.96 \pm 0.01$ & $0.96 \pm 0.01$ \\
    Superconductor  & $0.56 \pm 0.06$ & $0.56 \pm 0.06$ & $0.55 \pm 0.06$ & $0.55 \pm 0.06$  & $0.56 \pm 0.06$ & $0.57 \pm 0.05$ \\
    TFBind8 & $0.98 \pm 0.02$ & $0.92 \pm 0.03$ & $0.92 \pm 0.05$ & $0.93 \pm 0.04$  & $0.90 \pm 0.05$ & $0.88 \pm 0.07$ \\
    \bottomrule
  \end{tabular}
\end{table*}

\subsubsection{PGS performance w/ varying dataset (number of trajectories) size for offline RL}
Table \ref{tab:app_no_traj_offline} shows the ablation results for comparing PGS by training the policy with different dataset sizes (number of trajectories) using offline RL algorithm. Concretely, we ran CQL algorithm on three different sets of number of trajectories (1000, 10000, 20000) and evaluate the final performance. As evident from the table, the performance of PGS roughly increases as we increase the dataset size for offline RL training. 

\begin{table*}[h!]
  \caption{Table showing the ablation of different training dataset sizes (number of trajectories) for the offline RL algorithm in PGS.}
  \label{tab:app_no_traj_offline}
  \centering
  \begin{tabular}{llll}
    \toprule
    Number of trajectories  & 1k  & 10k & 20k  \\
    \midrule
    AntMorphology & $0.891 \pm  0.036$ & $0.889\pm 0.022$ & $0.949\pm 0.017$  \\
    DKittyMorphology & $0.943 \pm  0.010$ & $0.964\pm 0.014$ & $0.966\pm 0.013$  \\
    Superconductor  & $0.535 \pm  0.061$ & $0.521\pm 0.059$ & $0.563\pm 0.058$ \\
    TFBind8 & $0.829 \pm 0.022$ & $0.900\pm 0.080$ & $0.981\pm 0.015$ \\
    \bottomrule
  \end{tabular}
\end{table*}


\end{document}